\pdfoutput=1
\documentclass[11pt]{article}

\usepackage{acl}

\usepackage{times}
\usepackage{latexsym}

\usepackage[T1]{fontenc}
\usepackage[utf8]{inputenc}

\usepackage{microtype}
\usepackage{times}
\usepackage{url}
\usepackage{latexsym}
\usepackage{booktabs}

\usepackage{graphicx}

\usepackage{todonotes}
\usepackage{hyperref}

\title{MonoByte: A Pool of Monolingual Byte-level Language Models}

\author{Hugo Abonizio \\
  FEEC, UNICAMP, Brazil \\
  NeuralMind, Brazil \\
  {\tt hugo.abonizio@gmail.com} \\
  \And
  Leandro Rodrigues de Souza \\
  FEEC, UNICAMP, Brazil \\
  {\tt l231250@g.unicamp.br} \\
  \AND
  Roberto Lotufo \\
  FEEC, UNICAMP, Brazil \\
  NeuralMind, Brazil \\
  {\tt lotufo@unicamp.br } \\
  \And
  Rodrigo Nogueira \\
  FEEC, UNICAMP, Brazil \\
  NeuralMind, Brazil \\
  {\tt rfn@unicamp.br} \\
}

\date{}

\begin{document}
\maketitle
\begin{abstract}
  The zero-shot cross-lingual ability of models pretrained on multilingual and even monolingual corpora has spurred many hypotheses to explain this intriguing empirical result. However, due to the costs of pretraining, most research uses public models whose pretraining methodology, such as the choice of tokenization, corpus size, and computational budget, might differ drastically. When researchers pretrain their own models, they often do so under a constrained budget, and the resulting models might underperform significantly compared to SOTA models.
  These experimental differences led to various inconsistent conclusions about the nature of the cross-lingual ability of these models.
  To help further research on the topic, we released 10 monolingual byte-level models\footnote{\url{https://huggingface.co/monobyte}} rigorously pretrained under the same configuration with a large compute budget (equivalent to 420 days on a V100) and corpora that are 4 times larger than the original BERT's. Because they are tokenizer-free, the problem of unseen token embeddings is eliminated, thus allowing researchers to try a wider range of cross-lingual experiments in languages with different scripts.
  Additionally, we release two models pretrained on non-natural language texts that can be used in sanity-check experiments.
  Experiments on QA and NLI tasks show that our monolingual models achieve competitive performance to the multilingual one, and hence can be served to strengthen our understanding of cross-lingual transferability in language models.
\end{abstract}

\section{Introduction}

Shortly after the publication of BERT~\cite{devlin-etal-2019-bert}, researchers showed that multilingual models pretrained on the masked language modeling objective can achieve remarkable zero-shot cross-lingual performance on various NLP tasks (i.e., a multilingual model finetuned on a high-resource language and directly evaluated in other languages)~\cite{Conneau2020, xtreme_paper}. 

% For example, a model pretrained on a multilingual corpus and finetuned on a task that contains only English texts can perform well on that same task but on texts in other languages~\cite{Pires2019, xtreme_paper}.

These empirical results triggered a wave of research that aimed to explain this behavior. \citet{Pires2019} raised the ``anchor tokens'' hypothesis, i.e., that tokens shared between two languages act as a reference point so models can learn similar concepts. \citet{Artetxe2019} and \citet{Wu2020} questioned these findings and pointed out that, even without a shared vocabulary, models achieve great cross-lingual transfer performance by leveraging shared parameters only.

Perhaps even more surprising is the performance of \textit{monolingual} models in a cross-lingual setting. For instance, models pretrained and finetuned only on English can perform well in French tasks~\cite{jimmylin}. \citet{souza2021langagnostic} also show that monolingual models finetuned on a foreign language (i.e., a language that differs from pretraining) achieve comparable results to models finetuned on their language.

Recent work investigates properties of the pretraining corpora that contribute to the model's performance on natural language tasks. Models are pretrained on non-natural language corpora, such as code, music, proteins, and artificial languages and their performances are compared with natural language models~\cite{papadimitriou-jurafsky-2020-learning, pretrainwithouthuman, Lu2021, pretrainingartificial}. Evidence shows that pretraining on a corpus that contains artificial recursive or hierarchical structure between tokens results in a similar performance when compared to models pretrained on a natural language.

However, an outstanding problem is that these experiments are made either 1) using models pretrained by different research groups, whose pretraining configuration differs widely, or 2) using models pretrained under a compute budget or pretraining corpus that are orders of magnitude smaller than the ones used by SOTA models. For instance, \citet{pretrainwithouthuman} pretrain their models on a dataset 200 times smaller than BERT's. This is problematic because certain skills are learned only when corpus size and training budget are large enough~\cite{bowman}.

These differences in the pretraining methodology and undertrained models make it difficult to draw a conclusion about the nature of the cross-lingual ability of such models. Additionally, since most models use subword tokenization, it is difficult to experiment with languages that do not have a significant subword vocabulary overlap. For example, if a model is pretrained and finetuned on English, it cannot be tested in Chinese because there is very little token overlap in their vocabularies, and hence, Chinese token embeddings are not learned.
This is a problem even for languages with the same script. For example, many Spanish embeddings are not learned in an English-only pretraining, and hence it is difficult to tell whether a model's inability to learn cross-lingual representations is due only to tokenization issues~\cite{rust-etal-2021-good}.

Following \citet{foundation}, who advocate for the release of pretraining models, corpus, and script as a way to strengthen the field, we release byte-level models pretrained on large corpora from the same domain and using exactly the same training setup and compute budget.
Because they rely only on bytes to represent strings, they can be used to compare languages that use different scripts.

Each model takes approximately 210 hours of pretraining, resulting in more than three months of TPU computing. The models are available at \href{https://huggingface.co/monobyte}{https://huggingface.co/monobyte} and the code used for finetuning can be found at \href{https://github.com/lersouza/lang-agnostic}{https://github.com/lersouza/lang-agnostic}.

\section{Related work}

% \medskip
% \noindent\textbf{Monolingual models}. 
In the last years, many monolingual versions of BERT~\cite{devlin-etal-2019-bert, bertimbau, germanbert, camembert, arabert, korealbert, phobert, beto-model, banglabert} and T5~\cite{t5, ptt5, it5} have been released. The authors often claim that these versions outperform multilingual models in their languages. \citet{rust-etal-2021-good} showed that the tokenizer plays a critical role in achieving those results. More recently, \citet{xue2021byt5} released the ByT5 model, which would overcome this issue by leveraging a byte-level vocabulary. This model is pretrained on several languages and achieves great results when compared to mT5 \cite{xue-etal-2021-mt5}. However, no monolingual version of ByT5 has been released, which makes it difficult to conduct further investigation on the model's pretraining and cross-lingual performance that is not impacted by the tokenization in different languages.

% \medskip
% \noindent\textbf{Model pretraining research}.
Current research on the properties of pretraining often relies on monolingual models or models pretrained on artificial languages. Researchers often rely on models released by other groups~\cite{souza2021langagnostic} or pretrain models in a very controlled setup~\cite{Artetxe2019, papadimitriou-jurafsky-2020-learning, Fujinuma2022}. In both cases, it is not clear if the results suffer from these different methodologies or undertrained models. \citet{Blevins2022}, for instance, suggests that language contamination (i.e., a monolingual corpus composed of sentences in more languages) is the reason for monolingual zero-shot effectiveness. The same happens with models pretrained on non-natural language corpora~\cite{pretrainwithouthuman, Lu2021, pretrainingartificial}, where the corpus size differs from the ones trained on natural language, which might affect the conclusions drawn from the experiments. We hope to bridge this gap by releasing the set of monolingual models presented in this paper.

\section{Pretraining}

To train our monolingual models, we used mC4~\cite{xue-etal-2021-mt5}, the same corpus used to train mT5 and ByT5, which comprises 101 natural languages extracted from the Common Crawl web scrape. For each model, we selected only the documents written in the specific language, which was originally identified using \texttt{cld3} by \citet{xue-etal-2021-mt5}. The mC4 is a large corpus with different language distributions, with some being more represented than others. Thus, we trimmed each pretraining corpus in approximately 65 billion UTF-8 bytes for all languages.
The only exception was Bengali, which was trimmed on 32 billion bytes, corresponding to its total size.
% The only two exceptions were Bengali and Swahili, which were trimmed on 32 billion and 3 billion bytes, respectively, which are their total sizes.

The pretraining was conducted on a TPU VM v3-8 using Flax~\cite{flax2020github} library and the pretraining script for T5-like span-masked language modeling available on HuggingFace Transformers library.\footnote{\url{https://github.com/huggingface/transformers/blob/main/examples/flax/language-modeling/run_t5_mlm_flax.py}} We chose the smaller architecture of ByT5, with 300 million parameters, and used similar hyper-parameters as reported on ByT5~\cite{xue2021byt5} and mT5~\cite{xue-etal-2021-mt5} papers. We did not experiment with larger models due to their computational cost. We set the sequence length to 1024 tokens (UTF-8 bytes) and train for 1 million steps with a batch of $2^{16}$ bytes, resulting in approximately 65 billion bytes. Since we could only fit a batch of 8 examples per device, we used gradient accumulation of 128 steps to achieve a larger batch size. The learning rate starts at $10^{-3}$ and decays linearly throughout the pretraining.

In comparison with the original ByT5, our models are pretrained on a 16 times smaller corpus. However, our pretraining corpus is 4 times larger than BERT's, which was trained on 3,300M words~\cite{devlin-etal-2019-bert} --- i.e., approximately 16 billion characters, considering an average of 5 characters per word in English~\cite{bochkarev2015average}. In comparison with GPT-2, our corpus is 1.6 times larger~\cite{radford2019language}.

% Comparing with other monolingual models research, \citet{rust-etal-2021-good} pretrained for 900,000 steps using a batch size of 64 and a sequence length of 128 and the remaining 100,000 steps with 512, resulting in approximately 10 billion tokens. Thus, their experiments using the mBERT tokenizer, which has an average byte length per token of 7.3, resulted in 77GB of pretraining. \citet{pretrainingartificial} generated 12.8M sentences based on the average English sentence length, which is approximately 20 words~\cite{flesch1979write}, resulting in 1.2 billion characters or 1.2GB of artificial text.
Compared with other monolingual models research, \citet{rust-etal-2021-good} pretrained on a corpus of approximately 77GB, according to the reported number of steps and tokens per batch. \citet{pretrainingartificial} pretrained on a corpus of 1.2GB of artificially generated text.
A more extensive comparison to other work is presented in Table~\ref{tab:corpus-sizes} and demonstrates that our pretraining is comparable to SOTA monolingual models and much larger than corpora used in the majority of monolingual studies.

% However, each model takes approximately 210 hours of pretraining, resulting in more than 2,100 hours of TPU, or USD 18,480 (USD 8.8/hour). Although our monolingual models may not achieve the best performance when compared to their multilingual counterparts, our objective is to provide monolingual tokenizer-free language models to the research community...

% CamemBERT 138GB
% FinBERT 4.9 billion tokens
% BERTimbau 17GB
% X IndoBERT 220M words
% FlauBERT 71GB
% AraBERT 24GB
% BETO 3 bilion words (https://users.dcc.uchile.cl/~jperez/papers/pml4dc2020.pdf)
% KoreALBERT 43GB (4.4B words 18B characters)
% AraELECTRA 77GB or 8.8B words
% BanglaBERT 35GB (bangla ou bengali)
% PhoBERT Vietnamese 20GB

% \subsection{Languages}

\begin{table}
    %\small
    \centering\centering\resizebox{0.50\textwidth}{!}{
    \begin{tabular}{lp{1.8cm}r}
    \toprule
    \textbf{Model name} & \textbf{Language} & \textbf{Size} \\
    \midrule
    PTT5~\cite{ptt5}                  & pt & 15 \\
    BERT~\cite{devlin-etal-2019-bert} & en & 16 \\
    BERTimbau~\cite{bertimbau}        & pt & 17 \\
    PhoBERT~\cite{phobert}            & vi & 20 \\
    AraBERT~\cite{arabert}               & ar & 24 \\
    BanglaBERT~\cite{banglabert}        & bn & 35 \\
    KoreALBERT~\cite{korealbert}            & ko & 43 \\
    North ByT5~\cite{NorthT5}                   & no & 45 \\
    %BETO                  & es & \\
    CamemBERT~\cite{camembert}             & fr & 138 \\
    IT5~\cite{it5}                   & it & 215 \\
    \midrule
    \citealp{pretrainwithouthuman}  & en, artificial                    & 0.08 \\
    \citealp{papadimitriou-jurafsky-2020-learning}             & many, artif.                       & 0.5 \\
    \citealp{deshpande2021bert}     & many         & 0.5 \\
    \citealp{pretrainingartificial} & many, artif. & 1.2 \\
    \citealp{rust-etal-2021-good}              & many   & 77 \\
    \midrule
    MonoByte (ours, each model) & many & 65 \\
    \bottomrule
    \end{tabular}
    }
    \caption{Comparison of pretraining corpus sizes in GB of monolingual models. Models which does not explicitly report the corpus size were estimated based on the reported amount of tokens.}
    \label{tab:corpus-sizes}
\end{table}

\begin{table*}[h]
    \centering
    \small
    \begin{tabular}{lcccccccc}
        \toprule
        &                       \textbf{de} & \textbf{en} & \textbf{es} & \textbf{pt} & \textbf{ru} & \textbf{vi} & \textbf{zh} & \textbf{avg.} \\
        \midrule
        \textbf{Nonsense} & 33.33 & 33.55 & 33.27 & 50.00 & 33.33 & 33.57 & 34.19 & 35.89 \\
        \textbf{Hierarchical} & 66.01 & 69.32 & 68.04 & 65.65 & 62.12 & 64.49 & 62.83 & 65.49 \\
        \midrule
        \textbf{Monolingual}    &  70.85  & 76.76 &  75.07  &  83.40    & 67.39     & 70.37     &  69.91 & 73.39  \\
        \textbf{Multilingual}    & 76.02 &  80.55 &  78.39  &  67.73    &   73.58   &  73.01     &  73.39 & 74.67  \\
        \midrule
        $\Delta (\small{\textit{multi - mono}})$ & 5.16 & 3.80 & 3.32 & -15.67 & 6.19 &	2.64 &	3.48 & 1.28 \\
        \bottomrule
    \end{tabular}
    \caption{Results (accuracy) for Natural Language Inference in each language.}
    \label{table:results-nli}
    
\end{table*}

\begin{table*}
    \centering
    \small
    \begin{tabular}{lccccccc}
        \toprule
        &                       \textbf{ar} & \textbf{bn} & \textbf{en} & \textbf{ko} & \textbf{pt} & \textbf{ru} & \textbf{avg.} \\
        \midrule
                \textbf{Nonsense} & 0.22 & 0.00 & 0.45 & 0.00 & 0.26 & 0.27 & 0.20 \\
        \textbf{Hierarchical} & 67.23 & 0.38 & 28.03 & 9.99 & 3.63 & 32.85 & 23.68 \\
        \midrule
        \textbf{Monolingual}    & 78.17    & 53.89     &  60.15  & 51.17  & 64.88     & 58.26 & 61.09   \\
        \textbf{Multilingual}   & 81.81    & 70.28     &  68.51  & 54.70  & 42.37   & 72.97  & 65.11 \\ \midrule
        $\Delta (\small{\textit{multi - mono}})$ & 3.65 & 16.39 &	8.36 & 3.54 & -22.51 &	14.71 & 4.02 \\
        \bottomrule
    \end{tabular}
    \caption{Results (F1 Score) for Question Answering in each language.}
    \label{table:results-qa}
    
\end{table*}

% \subsection{Baselines}
% \noindent\textbf{Baselines}.
To sanity-check and validate the performance of our pretrained models, we compared their results with different baselines proposed in the literature. The first synthetic pretraining strategy, called nesting parentheses, was proposed by~\citet{papadimitriou-jurafsky-2020-learning}, which recursively generates random symbols respecting a hierarchical structure. This was proposed to evaluate how this recursion structure of the pretraining corpus transfers into modeling real languages. Their work showed that the models pretrained on the synthetic hierarchical corpus were able to predict human language far better than other baselines (e.g., random, music, and code), with a perplexity comparable to real languages.

% The first and more straightforward sanity check it to compare with models without pretraining, i.e., random initialized model weights. However, previous work have shown that models without pretraining tend to perform poorly on downstream tasks~\cite{CITAR}. Therefore, we also evaluated more strong pretraining strategies that use synthetic examples.

We used the code released by the authors, with the vocabulary being the 50,000 most frequent words based on the first one million documents of the Spanish mC4 corpus, and their sampling probability based on the Zipf distribution over the same Spanish corpus.\footnote{\url{https://github.com/toizzy/tilt-transfer/}} We generated examples until reaching the same size as our natural languages corpora sizes to make a fair comparison.

The second pretraining baseline was proposed by~\citet{krishna-etal-2021-pretraining-summarization}, dubbed nonsense, where the authors explored the knowledge transfer hypothesis of the downstream performance improvement of transfer learning. The authors pretrained on documents consisting of random n-grams and achieve similar performance on summarization over different datasets when compared to pretraining on real language. We generated examples using the original code\footnote{\url{https://github.com/acmi-lab/pretraining-with-nonsense/} \nopagebreak} to match the same corpus size of our natural language models.

\section{Experiments}

Our models are evaluated on two downstream tasks: Natural Language Inference (NLI) and Question Answering (QA). We use a similar setting as the \textit{In-language model}, described by \citet{xtreme_paper}, where the model is finetuned and evaluated in the subset of the task that corresponds to its pretraining language. The following subsections provide details about the datasets used and the results. The monolingual models are compared to the ByT5 Small checkpoint. The latter is expected to perform better since it is pretrained on a much larger set of a more diverse corpus, as investigated by \citet{Fujinuma2022}. For additional information  on the finetuning procedure, please refer to Appendix \ref{appendix:finetuning_details}.

% \subsection{Natural Language Inference}
\medskip
\noindent\textbf{Natural Language Inference}.
The XNLI\footnote{We use the machine-translated version of XNLI for training in different languages. Available at \href{https://dl.fbaipublicfiles.com/XNLI/XNLI-MT-1.0.zip}{https://dl.fbaipublicfiles.com/XNLI/XNLI-MT-1.0.zip}} dataset~\cite{conneau2018xnli} is used for finetuning and evaluating our German, English, Spanish, Russian, Vietnamese, and Chinese models. For the Portuguese model, we select the Recognizing Textual Entailment (RTE) task of the ASSIN2 dataset~\cite{assin2}. Performance is measured by using the accuracy metric.

% \medskip
Results are reported in Table \ref{table:results-nli}. Compared to the multilingual ByT5, our models achieve competitive performance. The difference in accuracy is about 1.28 on average. The results in Russian represent the largest gap (6.19 points), while those in Vietnamese represent the smallest (2.64 points). Nonsense achieves a near random performance, while Hierarchical approaches the models pretrained on natural language, as also evidenced by \citet{papadimitriou-jurafsky-2020-learning} and \citet{pretrainwithouthuman}.

% \subsection{Question Answering}
\medskip
\noindent\textbf{Question Answering}.
The gold passage version of the TydiQA~\cite{tydiqa} dataset (\textbf{TydiQA-GoldP}) is selected for finetuning and evaluating the Arabic, Bengali, English, Korean and Russian models. For finetuning the Portuguese model, we use the FaQuAD~\cite{faquad2019}. Performance is measured by the F1 Score metric as described by \citet{squad}.

Results are reported in Table~\ref{table:results-qa}. For Portuguese, our model outperforms the Multilingual checkpoint by 22.51 points, while, for Bengali, it stays 16.39 points behind. Our monolingual models stay, on average, 4 points behind the multilingual model released by Google. Looking at the difference between monolingual and multilingual models for every language, we see a higher variation. Nonsense and Hierarchical results are very distant. Question answering is a more difficult task~\cite{Vania2021} and a good benchmark for model quality. We hypothesize that this difference in task complexity requires more from the model. Pretraining with token structure emphasis only may not provide the required knowledge for good performance.

\section{Conclusion}

In this work, we introduced a pool of 10 tokenizer-free language models pretrained on large monolingual corpora. We demonstrated that our monolingual models can achieve competitive results to the multilingual ByT5, although having a smaller and less diverse pretraining --- single language, compared to all languages concatenated.

The main goal of our work was to release a set of monolingual models that carefully follow the same pretraining methodology to support the research of monolingual language models, pretraining properties, and cross-linguality. We hope that by releasing our models, we bridge this gap of having a more controlled and reproducible setup for rigorous experiments.

\section*{Acknowledgments}
This research was partially funded by grants 2020/09753-5 and 2022/01640-2 from Fundação de Amparo à Pesquisa do Estado de São Paulo (FAPESP).
We also would like to thank Google Cloud for credits to support this work.

\bibliographystyle{acl_natbib}
\bibliography{main}

\clearpage
\appendix

\section{Finetuning Details}
\label{appendix:finetuning_details}

The experiments were carried out in a single NVIDIA A100 80GB GPU. We use the Adafactor optimizer~\cite{adafactor} with a constant learning rate of $10^{-4}$. The models were finetuned using three different seeds, and we report an average of the results. We have chosen hyperparameters based on preliminary experiments with the ByT5 Small checkpoint\footnote{Available at: \url{https://huggingface.co/google/byt5-small}} on both XNLI and TydiQA datasets. We do not perform any exhaustive hyperparameter search, and use the same settings (per-task) for all models.

\medskip
\noindent\textbf{Natural Language Inference}. For all NLI experiments, we use a batch size of 16, accumulating gradients for 4 steps. The maximum input length was 1024, trimmed by batch. We train the model to output the class identifier (a number). We train for 10 epochs and evaluate every 0.2 epoch. We also perform early stopping with patience of 5 evaluations and select the best model on the validation set of each task. The results are reported against the test set of each dataset.

\medskip
\noindent\textbf{Question Answering}. For QA, the selected batch size is 6, accumulating gradients for 4 steps. The maximum input length was 2048 (question and context concatenated), trimmed by batch. We train the model to output an answer to the question with a maximum length of 768 bytes. We train for 10 epochs and evaluate at the end of each epoch. We also perform early stopping with patience of 3 and select the best model on the validation set of each task. 

\end{document}